\documentclass[11pt]{article}

\usepackage[utf8]{inputenc}
\usepackage[T1]{fontenc}
\usepackage[margin=1in]{geometry}

\usepackage{amsmath,amssymb}
\usepackage{nicefrac}

\usepackage{array}
\usepackage{booktabs}
\usepackage{multirow}
\usepackage{placeins}
\usepackage{pdflscape}
\usepackage{longtable}
\extrafloats{50}

\usepackage{graphicx}
\graphicspath{%
  {01_introduction/}%
  {03_methods/}%
  {05_geometry/}%
  {06_transformations/}%
  {06_abstract/}%
  {07_discussion/}%
  {08_conclusions/}%
  {figures/}%
  {tables/}%
}

\makeatletter
\def\input@path{%
  {01_introduction/}%
  {03_methods/}%
  {05_geometry/}%
  {06_transformations/}%
  {06_abstract/}%
  {07_discussion/}%
  {08_conclusions/}%
  {tables/}%
}
\makeatother

\usepackage{natbib}
\usepackage{xcolor}
\usepackage[colorlinks,citecolor=blue!60!black,linkcolor=blue!60!black,urlcolor=blue!60!black]{hyperref}

\title{Epicure: Navigating the Emergent Geometry\\
  of Food Ingredient Embeddings}
\author{
  Jakub Radzikowski\\
  KAIKAKU.AI\\
  \texttt{jakub@kaikaku.ai}
  \and
  Josef Chen\\
  KAIKAKU.AI\\
  \texttt{josef@kaikaku.ai}
}
\date{}

\begin{document}
\maketitle

\begin{abstract}
FlavorGraph~\citep{park2021flavorgraph} is the most comprehensive
public food embedding to date, combining FlavorDB chemistry with
Recipe1M+ co-occurrence into a single Metapath2Vec model.  In
earlier work we showed that FlavorGraph's 300-D embeddings already
encode at least fifteen interpretable culinary dimensions -- spanning
taste, texture, nutrition, geography, culture, and processing -- and
that LLM-augmented vocabulary consolidation strengthens most of those
signals~\citep{epicure2026}.  That study was tied to a single
English-centric pretraining, however, and fused chemical and
recipe-context signal as a fixed inductive bias rather than a
controllable design axis.  We present \textbf{Epicure}, a family of
three sibling skip-gram ingredient embeddings retrained from scratch
on a multilingual recipe corpus.

We aggregate 4.14M recipes from 11 sources spanning seven languages
(English, Chinese, Russian, Vietnamese, Spanish, Turkish, Indonesian,
German, and Indian-English) and normalise the raw ingredient strings to 1{,}790 canonical
entries via an LLM-augmented pipeline.  A 203{,}508-edge
ingredient--ingredient NPMI graph and an 80{,}019-edge typed FlavorDB
ingredient--compound graph (2{,}247 typed compound nodes across 15 categories)
seed three Metapath2Vec variants that share architecture and
hyperparameters and differ only in the random-walk schema:
\textbf{Cooc} walks the co-occurrence graph only, \textbf{Chem}
walks the typed compound metapaths only, and \textbf{Core} blends
both via injected ingredient--ingredient walks at controlled mixing,
placing each model at a distinct point on the
chemistry-vs-recipe-context spectrum.

All three Epicure models linearly recover supervised probes: 27
continuous sensory and nutrient directions and 8 cuisine
macro-regions, with mean Cohen's $d$ for cuisine separability of
$2.43$/$2.70$/$3.07$ for Cooc/Core/Chem.  An unsupervised
multi-seed-stable FastICA decomposition on food-group-residualised
embeddings recovers 20 interpretable factors per model, and a
Gaussian-mixture-model (GMM) partition of each factor's
high-quartile yields 150--200 named culinary modes per model with
mean coherence $0.611$/$0.833$/$0.703$ against random-pair baselines
of $0.097$/$0.348$/$0.115$.  Two complementary operator families run
on the same 300-D embedding: nearest-neighbour \emph{pairings}
(top-$K$ and mode-membership lookups) and SLERP \emph{direction
arithmetic} that rotates a seed toward either a supervised pole
vector (\emph{rice}~$+$~South-Asian $\to$ \emph{curry leaf, urad dal,
chana dal, fenugreek seed}) or an emergent factor-mode pole,
controlled by a continuous angle $\theta$ that interpolates between
seed-dominated and target-dominated retrieval.

The three sibling embeddings make chemistry-vs-recipe-context a
controllable design axis at the walk schema and expose both
label-grounded and emergent navigation operators on a single 300-D
space, supporting chef-facing tools that can rotate, blend, or
retrieve along either supervised semantic directions or
culturally-coherent emergent modes.  Code and trained artefacts are
not released at this time.

\end{abstract}

\section{Introduction}\label{sec:introduction}

A chef asked what pairs with \emph{miso} reaches for \emph{mirin},
\emph{dashi}, or \emph{sesame oil}.  Asked what pairs with
\emph{olive oil}, they reach for \emph{basil}, \emph{tomato}, or
\emph{prosciutto}.  Such choices are knowledge embedded in
recipe corpora across cultures and embodied in the working
intuition of cooks and chefs.  A computational representation of
this knowledge would enable a class of downstream tools: menu and
recipe assistants that surface plausible companions for an
ingredient on hand; cross-cuisine navigation that lets a
Mediterranean seed find its East-Asian peers without manual lookup; and
sensory- or nutrient-aware exploration that places an ingredient
inside an interpretable axis (fatty, fermented, bitter,
high-protein).  A useful ingredient embedding model is the substrate for all of these.

Computational gastronomy has approached this target from two
complementary directions.  \citet{ahn2011flavor} introduced the
\emph{flavour network} and established cultural divergence in
compound-sharing as an empirical phenomenon.
\citet{garg2018flavordb} catalogued the aroma molecules of 936 food
entities (FlavorDB), and \citet{foodb2020} extended chemical
coverage to FooDB's 70{,}000 compounds.  These chemical resources
underpin FlavorGraph~\citep{park2021flavorgraph}, which combined
FlavorDB with Recipe1M+~\citep{marin2021recipe1m,salvador2017learning}
in a heterogeneous graph of 6{,}653 ingredients and 1{,}645 compounds
trained with Metapath2Vec; it is the most comprehensive
public food embedding to date.  Symbolic alternatives such as
FoodKG~\citep{haussmann2019foodkg} integrate recipe, nutrition, and
ontology data into RDF knowledge graphs targeted at recommendation.

A separate line of work studies what kinds of computations a dense
embedding actually supports, and this paper draws directly on three
of its findings.  \citet{mikolov2013distributed} established that
semantic relationships emerge as linear directions in word2vec
(\emph{king}~$-$~\emph{man}~$+$~\emph{woman}~$=$~\emph{queen});
the directional view underwrites both our 27 supervised culinary
probes (cuisine, food-group, NOVA, USDA macronutrients, sensory)
and the 20 unsupervised FastICA factors we recover per model,
together with the SLERP rotation operator that traverses any of
them continuously.  \citet{mu2018allbutthetop} argued that
embedding isotropy is a precondition for stable directional
operations and proposed post-hoc rescue methods (all-but-the-top,
whitening) for collapsed geometries; we measure isotropy directly
via participation ratio and average pairwise cosine and find that
our three siblings sit at sharply different points on that
spectrum~-- a property of the walk schema rather than the input
data.  \citet{caliskan2017weat}'s Word Embedding Association Test
(WEAT) provides the standard diagnostic for whether named semantic
axes are reflected in the geometry; we report it in the supplement
alongside other robustness checks.

In earlier work~\citep{epicure2026} we analysed FlavorGraph's
300-D embeddings and found at least fifteen interpretable culinary
dimensions -- spanning taste, texture, nutrition, geography,
culture, and processing -- with LLM-augmented vocabulary
consolidation strengthening most of those signals.  That analysis
was bounded by FlavorGraph's fixed pretraining on three counts: a
single English-centric corpus, a single mix of chemistry and
recipe-context signal, and a scattered ingredient vocabulary that
included preparation details and non-food items.

We present \textbf{Epicure}, a family of three skip-gram ingredient
embeddings retrained from scratch to lift those three bounds
simultaneously.  We aggregate a 4.14M-recipe multi-language corpus
(English, Chinese, Russian, Vietnamese, Spanish, Turkish,
Indonesian, German, and Indian-English), normalise it to a shared
1{,}790-ingredient LLM-curated canonical vocabulary, and expose the
chemistry-vs-recipe-context mix as a controllable design axis at
the walk schema.  All three siblings share architecture and
hyperparameters; they differ only in which random walks the
skip-gram objective sees: \textbf{Cooc} walks recipe co-occurrence,
\textbf{Chem} walks typed FlavorDB compound--ingredient metapaths,
and \textbf{Core} blends both via injected ingredient--ingredient
walks at controlled mixing.  The three siblings thus trace the
chemistry-vs-recipe-context spectrum from a single experimental
design.

In the trained embeddings, supervised directions for cuisine,
food-group, NOVA processing class, USDA macronutrients, and 19
sensory categories are linearly recoverable; an unsupervised
multi-seed-stable FastICA decomposition on top of the embeddings
rediscovers 20~interpretable axes per model, and
Gaussian-mixture-model (GMM) partitioning of each factor's
high-quartile yields 150--200 named culinary modes
per model.  The supervised and emergent geometries together expose
two operator families on the same 300-D embedding: nearest-neighbour
\emph{pairings} (top-$K$ neighbours plus mode-membership lookup) and
SLERP \emph{direction arithmetic} that rotates a seed toward either
a supervised pole vector or an emergent factor-mode pole.

\section{Methods}\label{sec:methods}
The pipeline runs in five stages: (i) aggregate a multilingual recipe
corpus, (ii) normalise the raw NER terms into a canonical ingredient
vocabulary, (iii) construct the co-occurrence and typed-compound
graphs, (iv) train three Metapath2Vec variants (Cooc, Core, Chem) on
those graphs, and (v) analyse the resulting embeddings with
supervised direction probes and unsupervised factor / mode discovery.

\subsection{Corpus}\label{sec:data-sources}

We aggregate recipes from 11~publicly available datasets spanning
seven languages, yielding 4{,}135{,}189~recipes dominated by the
English RecipeNLG~\citep{bien2020recipenlg} (53.9\%) and the Chinese
XiaChuFang~\citep{liu2022xiachufang} (37.4\%) corpora, with the
Russian Povarenok~\citep{povarenok2021} corpus contributing 3.5\% and
eight smaller multilingual corpora covering
Vietnamese~\citep{anhnq1130cooking},
Spanish~\citep{somosnlprecetas,frorozcolrecetas,somosnlpabuela},
Turkish~\citep{sedatal2023turkish}, Indian (in English)~\citep{cleanedindianrecipes,indianfood101,southasianrecipes},
Indonesian~\citep{indonesianrecipes}, and German~\citep{chefkoch2021}.  Per-source recipe counts and macro-region
backing are catalogued in the supplement's \emph{Corpus and Vocabulary}
appendix.  Non-English
ingredient terms are machine-translated to English by the Claude Opus
family (internal deployment ID 4.6)~\citep{anthropic2026models}
under deterministic decoding (temperature~0); after merging,
deduplication, and intersecting with the final 1{,}790-canonical
vocabulary, 4{,}103{,}118~recipes (99.2\%) contain at least one
matched ingredient, with recipes carrying fewer than two matches
contributing no co-occurrence pair to the NPMI step.

\subsection{Canonical Vocabulary}\label{sec:normalisation}

Raw named-entity-recognition (NER) extraction across all eleven sources yields roughly
${\sim}200{,}000$ unique ingredient strings, dominated by spelling
variants, brand names, non-food items, and preparation modifiers.  An
LLM-augmented canonicalisation pipeline uses the Claude Opus family
(internal deployment ID 4.6)~\citep{anthropic2026models} with
deterministic decoding for term classification and Gemini Embedding
models for semantic clustering~\citep{lee2025geminiembedding}.
Production dedup runs used Google's API model identifier
\texttt{gemini-embedding-001}~\citep{google2026geminiembeddingapi},
followed by a final manual curation pass.  This reduces the set to
\textbf{1{,}790 canonical ingredients}.  Ingredient matching to
FlavorDB~\citep{garg2018flavordb} follows an entity-unique policy:
each FlavorDB entity matches at most one canonical ingredient, with
name-similarity tiebreaking when several candidates compete (the
supplement's \emph{Graph Construction} appendix details the policy).
After graph construction, 523 ingredients retain active typed I--C edges after the
\texttt{min\_compound\_degree=2} filter applied during graph
construction; the remaining 1{,}267 are non-hub.  Nutrient and sensory labels are matched
against USDA~FoodData Central~\citep{usda2019fooddata} and FlavorDB.  The
canonical-vocabulary CSV pairs each normalised ingredient name with its
FlavorDB and USDA anchors.  Principal
counted sets used throughout the paper are summarised in
Table~\ref{tab:vocab-chain}.

\begin{table}[!htbp]
\centering
\caption{Principal counted sets in the normalisation and evaluation
  pipeline.}
\label{tab:vocab-chain}
\small
\begin{tabular}{@{}lr@{}}
\toprule
Counted set & $n$ \\
\midrule
Final canonicals (embedded vocabulary)          & 1,790 \\
Cuisine-tagged total (pre co-occurrence filter) & 1,816 \\
\quad -- Universal (no distinctive region)              & 808 \\
\quad -- Cuisine-specific                                 & 1,008 \\
\qquad \textbullet~single-label                           & 858 \\
\qquad \textbullet~multi-label                            & 150 \\
Cuisine-clustering subset (specific $\cap$ embedded)   & 986 \\
Food-group-clustering subset (USDA $\cap$ embedded)    & 1,560 \\
\bottomrule
\end{tabular}

\end{table}

\paragraph{Cuisine taxonomy.}
For cuisine evaluation we define eight macro-regional cuisine clusters
grounded in corpus provenance (Table~\ref{tab:taxonomy}).  Claude Opus
family models (internal deployment ID 4.6)~\citep{anthropic2026models}
tag every canonical ingredient with zero or more macro-region labels
under a \emph{distinctive-marker} prompt: universal ingredients (salt,
onion, egg, flour, rice) are left untagged, and only ingredients that
immediately signal a culinary tradition receive a region label.
Of 1{,}816 tagged ingredients, 808 are universal (44.5\%)
and 1{,}008 are cuisine-specific (55.5\%); intersected with the final
1{,}790-canonical embedded set this yields 986 ingredients for
cuisine-clustering evaluation, of which 858 carry a single region label
and 150 carry two or three.

\begin{table}[!htbp]
\centering
\caption{Eight cuisine macro-regions and their approximate recipe-count
  backing in the training corpus.}
\label{tab:taxonomy}
\small
\begin{tabular}{@{}lrl@{}}
\toprule
Macro-region     & Backing recipes & Constituent traditions \\
\midrule
East Asian       & 1{,}549{,}034 & Chinese, Korean \\
Western Atlantic &    198{,}086 & American, British, German, Scandinavian \\
Mediterranean    &    164{,}107 & Italian, French, Iberian, Greek, Levantine, North African, Turkish \\
Eastern European &    154{,}479 & Russian, Ukrainian, Polish, Hungarian, Georgian \\
Southeast Asian  &    107{,}964 & Thai, Vietnamese, Filipino, Indonesian, Malay \\
South Asian      &     47{,}462 & Indian, Pakistani, Sri Lankan, Bangladeshi \\
Latin American   &     40{,}618 & Mexican, Caribbean, Brazilian, Peruvian, Colombian \\
Japanese         &     33{,}923 & Japanese \\
\bottomrule
\end{tabular}
\end{table}

\subsection{Graph Construction}\label{sec:graph-construction}

The three Epicure models share the same 1{,}790-ingredient node set and
the same 203{,}508 NPMI co-occurrence edges
(Table~\ref{tab:graph-stats}).  Two graph variants are constructed:

\paragraph{Cooc graph (co-occurrence only).}
Ingredient--ingredient edges weighted by normalised pointwise mutual
information~\citep{bouma2009normalized} (NPMI) computed over the
4.10M matched recipes.  Ingredients appearing in fewer than 20
recipes are dropped before NPMI computation, which together with the
canonicalisation pipeline yields the 1{,}790-ingredient vocabulary.
After retaining only positive-NPMI pairs, the graph has
203{,}508~edges.

\paragraph{Core/Chem graph (co-occurrence + typed compound edges).}
Adds 2{,}247 typed FlavorDB compound nodes connected to ingredient
nodes by 80{,}019 typed I--C edges.  Each original compound carries one or more of 15
flavor-category tags (balsamic, citrus, earthy, fatty, floral, fruity,
green, meaty, minty, nutty, spicy, vegetable, wine-like, woody, plus
one residual); compounds are replicated once per category they belong
to so Metapath2Vec's typed walks can distinguish a citrus--citrus
compound overlap from a citrus--earthy bridge.  This approach differs from FlavorGraph's single-type compound node, which is a single node for all compounds of a given type.

\begin{table}[!htbp]
\centering
\caption{Graph variant statistics.  Cooc operates on a pure
  ingredient--ingredient graph; Core and Chem share an identical
  heterogeneous graph that adds 2{,}247 typed FlavorDB compound
  nodes and 80{,}019 typed I--C edges.  All three models share the
  same 1{,}790-ingredient vocabulary and 203{,}508 NPMI
  co-occurrence edges; the difference between Core and Chem is the
  walk schema, not the graph.}
\label{tab:graph-stats}
\small
\begin{tabular}{@{}lrrr@{}}
\toprule
& Cooc graph & \multicolumn{2}{c}{Core/Chem graph} \\
\cmidrule(lr){3-4}
& (Cooc) & (Core) & (Chem) \\
\midrule
Ingredient nodes              & 1{,}790   & 1{,}790   & 1{,}790   \\
Compound nodes (typed)        & ---       & 2{,}247   & 2{,}247   \\
I--I edges (NPMI $> 0$)       & 203{,}508 & 203{,}508 & 203{,}508 \\
I--C edges (typed)            & ---       & 80{,}019  & 80{,}019  \\
Total typed graph edges       & 203{,}508 & 283{,}527 & 283{,}527 \\
Compound types                & ---       & 15        & 15        \\
Ingredients with I--C edges   & ---       & 523       & 523       \\
\bottomrule
\end{tabular}

\end{table}

\subsection{The Three Epicure Models}\label{sec:models}

We train three metapath2vec~\citep{dong2017metapath2vec} models with
identical architecture and hyperparameters
(Table~\ref{tab:training-hyperparameters}: 300-dim embeddings,
\texttt{walks\_per\_node}$=$100, \texttt{walk\_length}$=$50,
\texttt{context\_size}$=$7, 5 negative samples,
\texttt{batch\_size}$=$32{,}768, \texttt{lr}$=$0.0025, 20 epochs, no
warm restart).

The objective is skip-gram with negative
sampling~\citep{mikolov2013distributed}.

Implementation uses the PyTorch framework~\citep{paszke2019pytorch}.
We refer to the family collectively
as \textbf{Epicure} and to its three siblings as Epicure-Cooc,
Epicure-Core, and Epicure-Chem.  They differ only in which random
walks the skip-gram objective sees:

\begin{description}
\item[\textbf{Epicure-Cooc.}] Walks the Cooc graph: pure I--I random
  walks weighted by NPMI.  No compound nodes.
\item[\textbf{Epicure-Core.}] Walks the typed-compound graph and
  injects pure I--I walks at \texttt{--ii\_repeat=10} alongside the
  typed-compound metapaths.  Edge transitions are weighted so I--C
  hops are not oversampled relative to the smaller I--I edge set.  The
  resulting embedding blends chemical and recipe-context signal.
\item[\textbf{Epicure-Chem.}] Walks the typed-compound graph but with
  \texttt{--ii\_repeat=0}: the I--I templates are absent and the only
  walks the skip-gram sees are compound-mediated.  The chemistry
  extreme of the family.
\end{description}

\begin{table}[!htbp]
\centering
\caption{Metapath2Vec training hyperparameters.  All three Epicure
  models share every architecture and optimiser setting; the bottom
  block lists the walk-design choices that differ across the three
  variants.  All runs use PyTorch with SparseAdam and a fixed seed.}
\label{tab:training-hyperparameters}
\small
\begin{tabular}{@{}lccc@{}}
\toprule
Parameter & Cooc & Core & Chem \\
\midrule
Embedding dimension      & 300       & 300       & 300       \\
Walks per node           & 100       & 100       & 100       \\
Walk length              & 50        & 50        & 50        \\
Context window           & 7         & 7         & 7         \\
Negative samples         & 5         & 5         & 5         \\
Batch size               & 32{,}768  & 32{,}768  & 32{,}768  \\
Learning rate            & 0.0025    & 0.0025    & 0.0025    \\
Optimiser                & SparseAdam & SparseAdam & SparseAdam \\
LR schedule              & constant  & constant  & constant  \\
Epochs                   & 20        & 20        & 20        \\
Random seed              & 42        & 42        & 42        \\
\midrule
\multicolumn{4}{@{}l}{\textit{Walk design}} \\
Graph backbone           & Cooc      & \multicolumn{2}{c}{Core/Chem (typed)} \\
Typed compound walks     & ---       & yes       & yes       \\
Weighted I--C transitions & ---      & yes       & yes       \\
I--I walk injection (\texttt{ii\_repeat}) & native (only I--I) & 10$\times$ & 0 (none) \\
\bottomrule
\end{tabular}
\end{table}

The three models trace a chemistry-vs-recipe-context walk-template
spectrum from a single experimental design. Section~\ref{sec:geometry} characterises how this spectrum
manifests in the trained embeddings; Section~\ref{sec:transformations}
exploits it.

\paragraph{Walk metapaths in detail.}
Compounds attach only to the 523 ingredients with active I--C edges
(\S\ref{sec:normalisation}); following the FlavorGraph nomenclature
\citep{park2021flavorgraph}, we call these
\emph{chemical-hub} (\texttt{H}) ingredients and the remaining 1{,}267
\emph{non-hub} (\texttt{N}) ingredients.  With \texttt{C[x]} denoting a
compound of family $x$, Core and Chem generate three families of
typed-compound walks, each playing a distinct role:
\emph{within-type} \texttt{H--C[x]--H} aggregates ingredient pairs that
share a same-family compound;
\emph{via-compound} \texttt{N--H--C[x]--H--N} is the only route by
which non-hub ingredients receive compound-mediated context; and
\emph{cross-type} \texttt{C[x]--H--N--H--C[y]} bridges two compound
families through a hub--non-hub ingredient chain.  Each of the 15
compound types receives one within-type and one via-compound
template; $2n=30$ cross-type templates are sampled per walk round
with a coverage guarantee that every type appears as both source and
target.  Core additionally samples ten pure I--I templates per walk
round, so the I--I context is roughly an order of magnitude more
frequent than any single compound-mediated template.  We cycle
templates with naive \texttt{pos\,\%\,len(template)}, which
deviates from FlavorGraph's palindromic convention and concentrates
the chemistry signal into short, high-information walks; an
ablation in the supplement's \emph{Walk schema cycling} subsection
documents the resulting walk-length distribution and a side-by-side
comparison against the palindromic alternative.

\subsection{Evaluation}\label{sec:eval}

The trained embeddings are evaluated under three blocks that map
1:1 onto the geometry section that follows.

\paragraph{Direction quality.}
We score 27 continuous probes and 8 cuisine macro-regions, all
intersected with the three models' shared vocabulary, under 5-fold
repeated cross-validation.  Continuous probes report Spearman~$\rho$
between an ingredient's projection onto a fold-trained linear
direction and its ground-truth score; cuisine probes report
one-vs-rest Cohen's~$d$ on the distinctive-marker tags.  The
continuous probes are organised into three strata that
progressively decouple from the typed I--C walk schema: 14 baked-in
compound-feature (CF) sensory categories the schema sees directly
(e.g.\ \textsf{cf\_citrus}), 5 held-out basic-taste CF probes the
schema does not see, and 8 USDA macronutrient probes drawn from
external nutrient data (e.g.\ \textsf{usda\_protein\_g}); the 8
cuisine macro-regions, drawn from LLM-annotated distinctive-marker
tags (e.g.\ \emph{Japanese}), form a fourth stratum further removed
from the training signal.  Stratum design and the regression
protocol are detailed in the supplement's \emph{Stratified
Direction Quality} appendix.

\paragraph{Intrinsic geometry.}
Participation ratio (PR) and average pairwise cosine quantify
isotropy.  Normalised mutual information (NMI) measures
self-organisation around 17 USDA food groups (single-label) and 8
cuisine macro-regions (multi-label); the soft-NMI variant used for
the cuisine case is defined in the supplement's \emph{Multi-label
NMI protocols} subsection.  Silhouette and $k$NN@5 purity are
reported as auxiliary cluster-quality metrics in
Table~\ref{tab:intrinsic-metrics}.

\paragraph{Emergent geometry.}
20 ICA factors are extracted per model with
\texttt{sklearn.FastICA}~\citep{pedregosa2011sklearn} on the
\emph{food-group-residualised} embedding so the recovered axes are
orthogonal to the dominant food-group variance.  Factor
identifiability is enforced via Hungarian matching across 10
random seeds; the seed whose components have the highest mean
matched-cosine across the others is retained, factors are sorted
by stability descending (so factor index 0 is the most
reproducible), and only factors with split-half cosine stability
above $0.6$ are kept (supplement's \emph{Multi-seed FastICA
protocol} subsection).  For each ICA factor, the top-quartile
ingredients are partitioned in PCA-reduced space into
Gaussian-mixture-model modes under BIC over $K \in \{3,\dots,7\}$
with a six-member minimum per mode; each resulting mode is
projected back to 300-D as a unit-mean ``pole''.  The same GMM
procedure is run in parallel on the high-quartile of every
property in a curated supervised set (NOVA processing level,
CF/USDA/LLM sensory scores, food-group binaries) so that emergent
factor modes and supervised-property modes share the same
representation.  Mode coherence is the mean within-mode pairwise
cosine, baselined against random-pair samples of the same size.

\FloatBarrier

\section{Geometry}\label{sec:geometry}

We characterise the three Epicure embeddings in three steps:
isotropy and food-group separation (Section~\ref{sec:isotropy-foodgroup})
quantify how broadly each model spreads variance and how cleanly food
groups separate; supervised direction quality
(Section~\ref{sec:direction-quality}) measures how well linear
directions recover labelled probes; emergent geometry
(Section~\ref{sec:emergent}) reports the unsupervised ICA-$n{=}20$
factor analysis and the GMM modes that fall out of it, plus a
coherence metric quantifying how tight the modes are.  The 20~ICA
factors and 150--200 modes per model are the geometric vocabulary
that the operators in Section~\ref{sec:transformations} act on.

\subsection{Isotropy and food-group structure}\label{sec:isotropy-foodgroup}

\begin{table}[!htbp]
\centering
\caption{Intrinsic geometry of the three Epicure models.  Higher
  participation ratio (PR) and lower average pairwise cosine   indicate a more isotropic embedding.  Cuisine NMI is the soft
  multi-label variant.  Clustering metrics use the common
  labelled subset ($n{=}1{,}560$ food-group, $n{=}986$ cuisine);
  bracketed values are 95\% CIs from 0.8$n$ subsample bootstrap,
  200 iterations.  \textbf{Bold} marks the best result per row
  across the three models.}
\label{tab:intrinsic-metrics}
\small
\setlength{\tabcolsep}{4pt}
\begin{tabular}{@{}llccc@{}}
\toprule
Category & Metric & Cooc & Core & Chem \\
\midrule
\multicolumn{5}{@{}l}{\textit{Isotropy}} \\
& $N$ ingredients               & 1,790 & 1,790 & 1,790 \\
& Participation ratio $\uparrow$ & 173.6 & 94.2 & \textbf{183.1} \\
& Avg.\ pairwise cosine $\downarrow$ & \textbf{0.099} & 0.349 & 0.117 \\
& PCA top-10 variance & \textbf{0.138} & 0.234 & 0.141 \\
& PCA top-50 variance & \textbf{0.301} & 0.434 & 0.360 \\
\midrule
\multicolumn{5}{@{}l}{\textit{Food-group clustering (17 categories, $n{=}1{,}560$)}} \\
& NMI & 0.205\,[0.191, 0.223] & \textbf{0.235}\,[0.212, 0.251] & 0.226\,[0.215, 0.248] \\
& $k$NN@5 purity & 0.307\,[0.282, 0.307] & 0.352\,[0.328, 0.349] & \textbf{0.355}\,[0.332, 0.353] \\
& Silhouette & -0.036\,[-0.045, -0.028] & -0.053\,[-0.062, -0.037] & \textbf{-0.028}\,[-0.038, -0.015] \\
\midrule
\multicolumn{5}{@{}l}{\textit{Cuisine-region clustering (8 macro-regions, $n{=}986$)}} \\
& Soft NMI & \textbf{0.457}\,[0.402, 0.481] & 0.456\,[0.441, 0.509] & 0.432\,[0.399, 0.479] \\
& $k$NN@5 Jaccard purity & 0.652\,[0.630, 0.656] & \textbf{0.695}\,[0.670, 0.697] & 0.677\,[0.654, 0.680] \\
& Silhouette & +0.028\,[0.026, 0.030] & \textbf{+0.050}\,[0.045, 0.054] & +0.039\,[0.036, 0.041] \\
\bottomrule
\end{tabular}
\end{table}

In order to characterise the basic conditioning of each embedding
before testing linear operators on it, we measured two intrinsic
geometry diagnostics (participation ratio, average pairwise cosine)
and two unsupervised label-recovery diagnostics (normalised mutual
information against the 17 USDA-derived food groups and against the
eight cuisine macro-regions).

We found two isotropic geometries and one concentrated one: Cooc
reaches participation ratio $\mathrm{PR} = 173.6$ of 300 possible
dimensions and Chem $\mathrm{PR} = 183.1$, both with average pairwise
cosine in the $0.10$--$0.12$ band, while Core sits at $\mathrm{PR} =
94.2$ with average pairwise cosine $0.35$.  This means the
concentration in Core is a property of its walk schema rather than
its inputs: Core injects each ingredient--ingredient edge as a
length-2 walk and repeats those injected walks ten times per round
(Section~\ref{sec:methods}), creating strong recipe-context
attractors; Cooc lacks the typed I--C metapaths and Chem lacks the
injected I--I repetition, so both end up similarly spread.

\begin{figure}[!htbp]
\centering
\includegraphics[width=\textwidth]{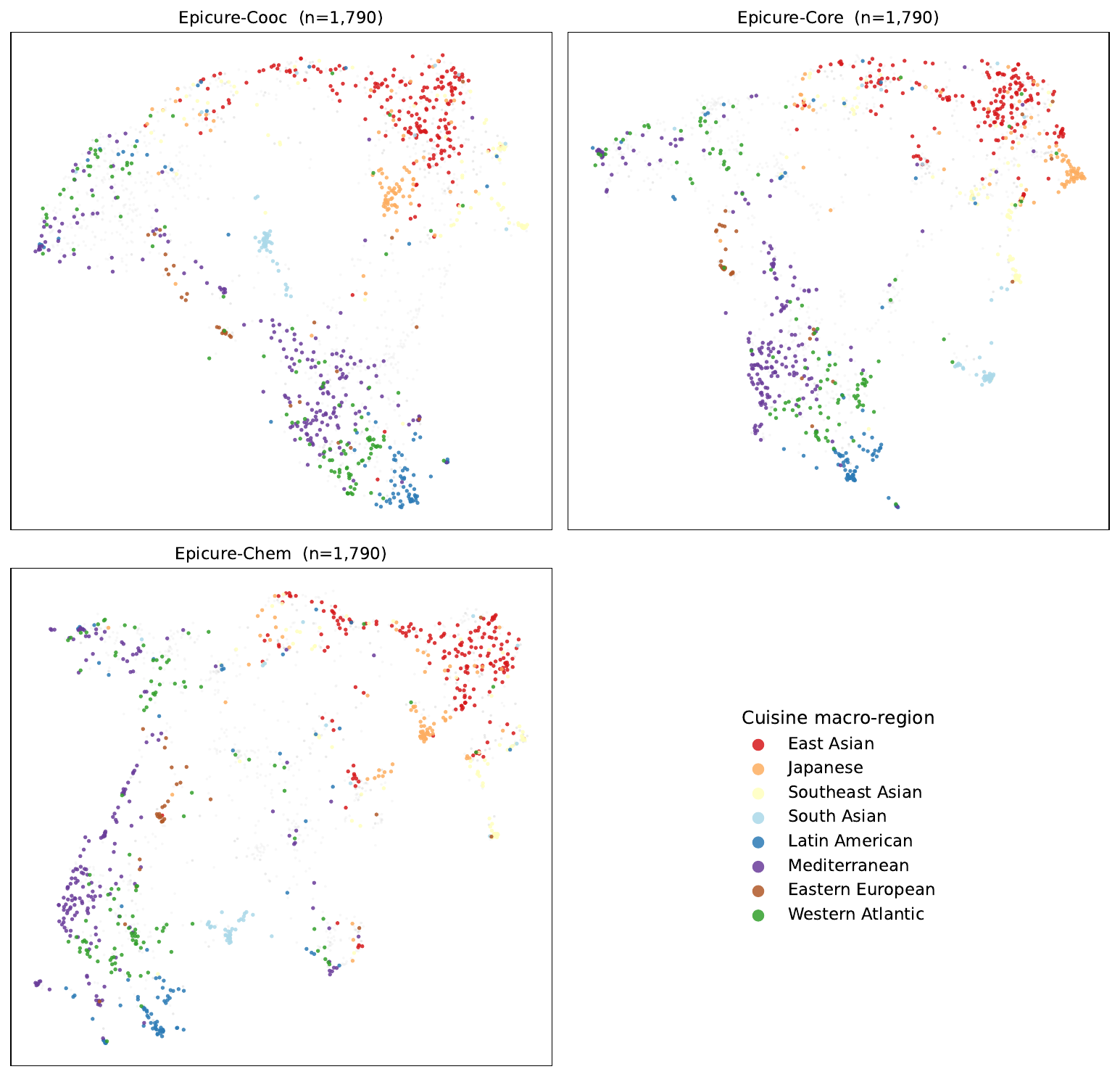}
\caption{2-D UMAP projection (cosine, \texttt{n\_neighbors}${=}30$,
  \texttt{min\_dist}${=}0.03$) of each Epicure model's 1{,}790
  ingredients, coloured by cuisine macro-region; universally tagged
  ingredients are de-emphasised in grey so the cultural structure
  dominates visually.  All three models exhibit clearly separated
  East Asian, South Asian, Latin American, and Mediterranean
  clusters, with the tightness of those regions paralleling the
  isotropy ordering: Core's compressed geometry compresses the
  clusters as well, while the isotropic Cooc and Chem produce more
  diffuse but still cleanly partitioned regions.  The same UMAP
  coordinates, coloured by USDA food group, are reproduced in the
  supplement's \emph{UMAP Visualisations} appendix.}
\label{fig:cuisine-umap}
\end{figure}

We also found that all three embeddings organise themselves around
nutritional and cultural labels without those labels being used for
training: ingredients from the same USDA food group land closer
together than chance, scoring $0.20$--$0.25$ on normalised mutual
information (NMI; 0 = chance, 1 = perfect recovery), and soft NMI on
the eight cuisine macro-regions rises to $0.43$--$0.46$ across the
three models -- roughly double the food-group level.  This means
cultural tradition shapes ingredient co-occurrence more cleanly than
nutritional category.  Figure~\ref{fig:cuisine-umap} visualises the
cuisine structure: a 2-D UMAP projection of each model coloured by
cuisine macro-region surfaces visibly distinct East Asian, South
Asian, Latin American, and Mediterranean clusters in all three
Epicure variants.

Both label structures appear without any supervision; whether they
translate into usable linear directions is the next question.

\FloatBarrier

\subsection{Direction quality}\label{sec:direction-quality}

In order to test whether labelled culinary concepts are linearly
recoverable in each embedding -- and how that recoverability varies
as the probe decouples from the training signal -- we ran the
five-fold cross-validated direction-quality protocol of
Section~\ref{sec:eval} on the four-stratum probe set (14 baked-in CF
+ 5 held-out basic-taste CF + 8 USDA macros + 8 cuisine
macro-regions).

We found that all three models recover every stratum linearly, with
the same ordering Cooc $<$ Core $<$ Chem at each one: baked-in CF
$\bar{\rho} = 0.28 / 0.40 / 0.46$; held-out basic-taste CF $0.32 /
0.42 / 0.47$; USDA macros $0.41 / 0.45 / 0.49$; cuisine
macro-regions $\bar{d} = 2.43 / 2.70 / 3.07$.  Across the 27
continuous probes Chem beats Core on 26 and Cooc on 27, and leads on
8 of 8 cuisine regions.  This means linear directions are usable
navigation primitives in all three siblings, and the chemistry-heavy
walk schema (Chem) sharpens them most -- complementing rather than
overriding the recipe-context signal.  The supplement's
\emph{Stratified Direction Quality} appendix reports stratum-level
robustness checks, including an orthogonal-residual SNR ranking,
$\ell_1$-regularised linear probes on categorical and continuous
targets, and a held-out cross-modal validation against external
FlavorDB and USDA labels.

Supervised directions answer ``where labelled concepts live''; the
embedding's own natural axes need not coincide with any label, which
the next subsection takes up.

\begin{figure}[!htbp]
\centering
\includegraphics[width=\textwidth]{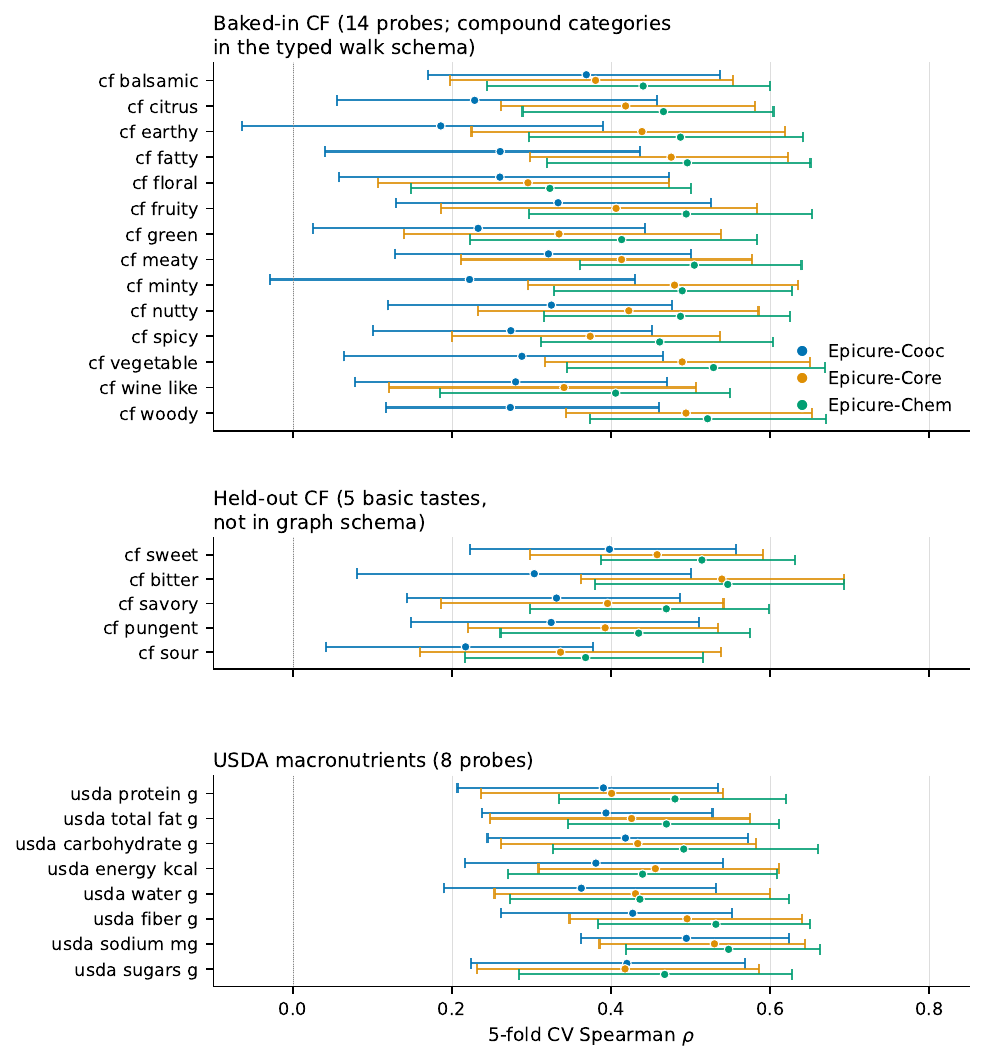}
\caption{Direction quality as 5-fold repeated cross-validated
  Spearman~$\rho$ between each ingredient's projection onto the
  linear direction (positive vs.\ negative pole separation) and its
  ground-truth score, with point estimate and 95\% CI per Epicure
  model.  The 27 continuous probes split into three strata: 14
  FlavorDB compound-feature (CF) sensory categories whose labels
  index Core's and Chem's typed I--C walk schema (e.g.\
  \textsf{cf\_citrus}); 5 basic-taste CF probes outside the graph
  schema; and 8 USDA macronutrient probes from external nutrient
  data.  Chem (green) leads on every probe except
  \textsf{usda\_energy\_kcal}, with the consistent ordering Cooc $<$
  Core $<$ Chem across rows.}
\label{fig:direction-quality}
\end{figure}

\begin{figure}[!htbp]
\centering
\includegraphics[width=\textwidth]{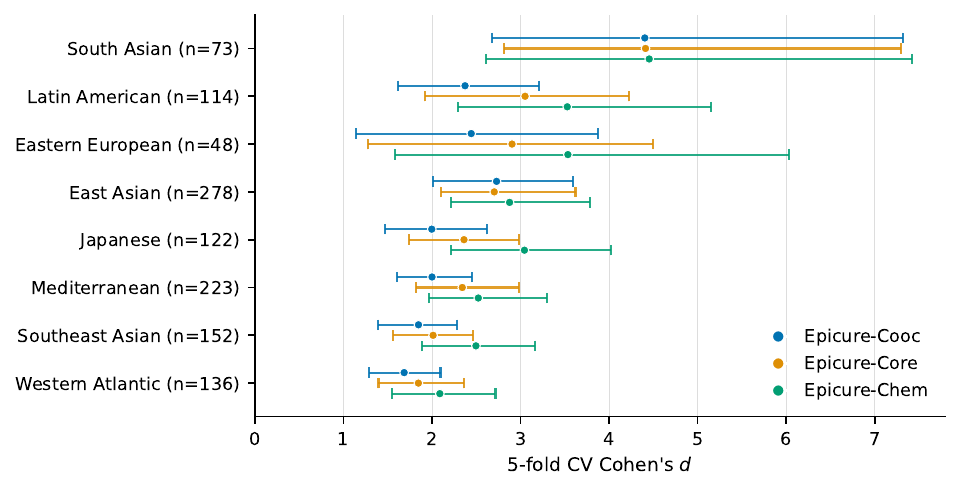}
\caption{Per-region Cohen's~$d$ (5-fold repeated CV, one-vs-rest on
  the distinctive-marker tags for each macro-region) for the three
  Epicure models, with 95\% CIs.  $n$ is the number of tagged
  ingredients per region; higher $d$ means more linearly separable.
  Regions are sorted by mean $d$ across models.  Chem leads on 8 of
  8 regions; CIs widen sharply for low-$n$ regions (Eastern
  European, South Asian) but the cross-region ranking is consistent.}
\label{fig:cuisine-d}
\end{figure}

\FloatBarrier

\subsection{Emergent factors and modes}\label{sec:emergent}

In order to discover the embedding's natural axes without using any
labels, we ran the multi-seed-stable FastICA + GMM mode-discovery
pipeline of Section~\ref{sec:eval} on each Epicure model; the
supplement's \emph{Factor Decomposition} appendix documents the
factor-extraction method comparison, per-factor split-half
stability, and a cuisine-orthogonalisation robustness check.

We found 20 stable factors per model and 150--200 modes per model
(Cooc 150 modes across 41 properties; Core 193 / 44; Chem 200 / 43),
each reading as a named culinary neighbourhood: \emph{Sweet baking
and dessert ingredients}, \emph{South Asian whole spice blends},
\emph{Mexican \& Latin American Pantry}.  Per-model mode listings
are in the supplement's \emph{Mode Atlas} appendix.
Figure~\ref{fig:mode-umap-3panel} renders one representative factor
per model as a worked example: the top-quartile of each factor is
coloured by GMM-mode assignment on the model's 2-D UMAP, with a
short Claude-generated label at every mode's centroid.  We discuss
factor indices as coordinates that locate modes rather than as named
axes themselves; the interpretable culinary content lives at the
mode level, not the factor pole.

\begin{figure}[!htbp]
\centering
\includegraphics[width=\textwidth]{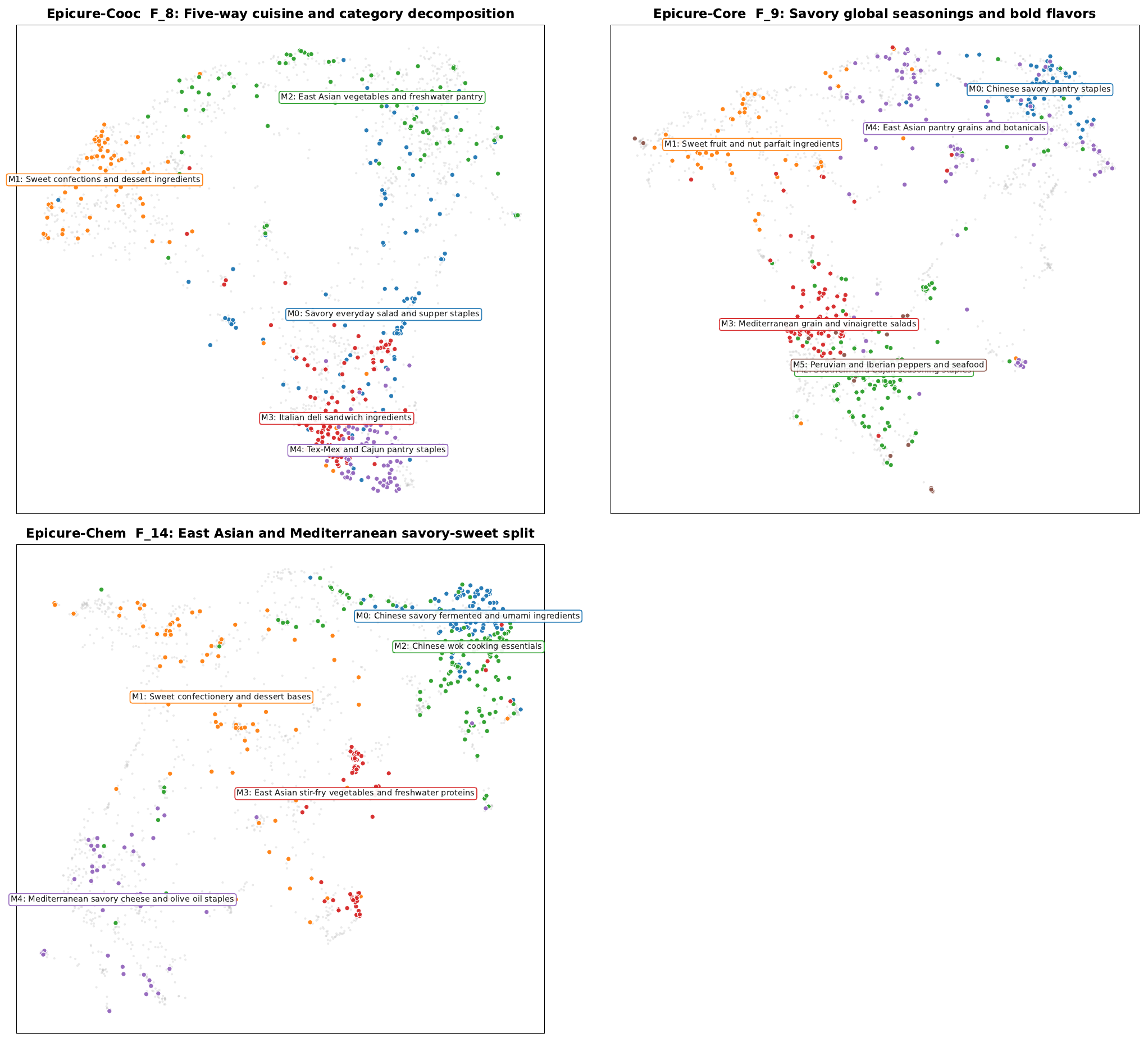}
\caption{One ICA factor per Epicure model with its GMM-mode
  decomposition, projected onto each model's own 2-D UMAP.  Coloured
  points are the top-quartile members of the highlighted factor,
  partitioned into GMM modes with Claude-generated labels at each
  mode's median centroid; grey points are the rest of the
  1{,}790-ingredient vocabulary.  Each panel title carries a short
  Claude summary of the factor's high-quartile derived from its K
  mode labels: a single named culinary identity when the modes
  cohere, or a description of the multi-cluster decomposition
  itself when they do not (here, all three picks fall in the
  decomposition regime -- Cooc $F_8$ splits into five distinct
  cuisine families, Core $F_9$ into six savoury seasoning
  sub-clusters, Chem $F_{14}$ into an East-Asian vs.\
  Mediterranean savoury--sweet split, demonstrating that
  even compound-mediated metapaths surface multi-modal culinary
  geometry).  ICA orientations are model-specific so the three
  factor indices do not correspond across panels.  Full per-model
  factor summaries and per-mode atlases are in the supplement.}
\label{fig:mode-umap-3panel}
\end{figure}

\begin{figure}[!htbp]
\centering
\includegraphics[width=\textwidth]{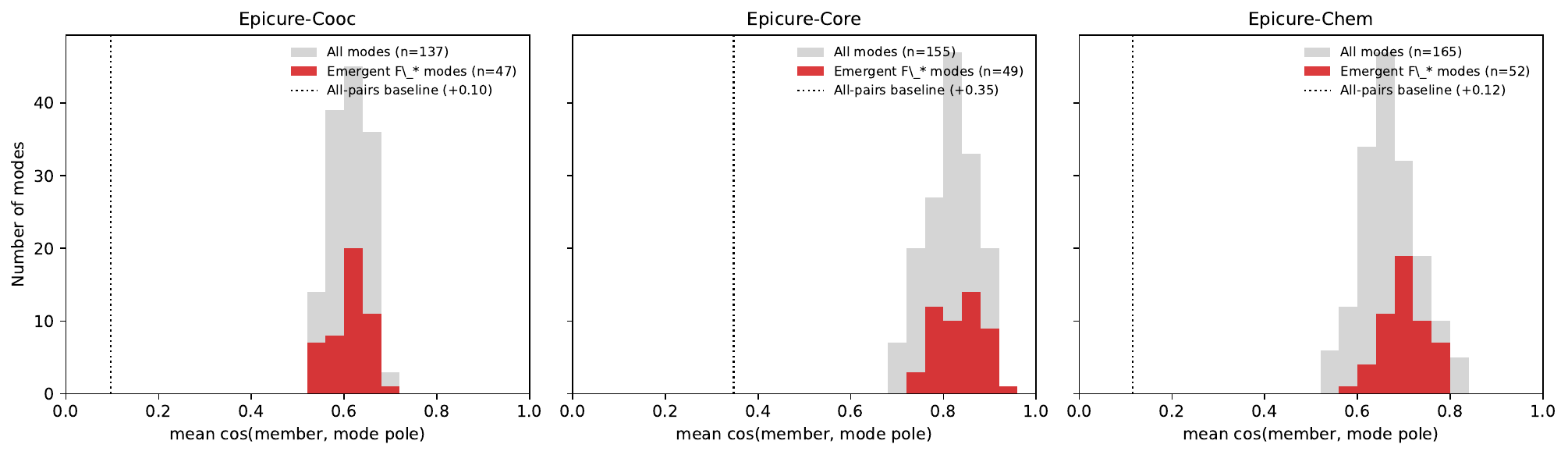}
\caption{Distribution of per-mode coherence (mean cosine of members
  to mode pole) for each Epicure model.  Red bars are emergent
  ($F_*$) modes; gray bars are all modes including supervised
  properties for context; dotted line is the all-pairs random-pair
  baseline.  Emergent modes sit well above baseline in every model:
  Cooc $0.611$ vs.\ baseline $0.097$, Core $0.833$ vs.\ $0.348$,
  Chem $0.703$ vs.\ $0.115$.  The tightness margin
  (mode-coherence~$-$~baseline) is comparable across models
  ($\approx 0.5$); absolute coherence tracks each model's overall
  concentration (Core's concentrated geometry pulls both pole
  tightness and the all-pairs baseline up; the isotropic Cooc and
  Chem sit lower in absolute terms with the same margin).}
\label{fig:mode-coherence}
\end{figure}

We also found that emergent modes sit 5--6$\times$ above the
random-pair coherence baseline in every model:
Figure~\ref{fig:mode-coherence} reports mean-cosine-to-pole of
$0.611$/$0.833$/$0.703$ for Cooc/Core/Chem against random-pair
baselines of $0.097$/$0.348$/$0.115$.  The tightness margin
(coherence~$-$~baseline) is comparable across the three models,
$\approx 0.5$; absolute coherence tracks each model's overall
concentration -- Core's $\mathrm{PR} = 94$ pulls both pole tightness
and the all-pairs floor upward, while the isotropic Cooc and Chem
($\mathrm{PR} \approx 174$ and $183$) produce lower absolute
coherence with the same margin.  This means the unsupervised axes
are not artefacts of a single seed and the modes that fall out of
them are tight named neighbourhoods rather than arbitrary partitions
-- a vocabulary of navigation atoms alongside the supervised
directions of Section~\ref{sec:direction-quality}.
Section~\ref{sec:transformations} demonstrates the operators that
act on these atoms.

\FloatBarrier

\section{Transformations}\label{sec:transformations}

The geometry of Section~\ref{sec:geometry} -- linearly recoverable
supervised directions plus 150--200 named emergent modes per model --
exposes two complementary operator families: nearest-neighbour
\emph{pairings} (Section~\ref{sec:pairings}) and SLERP-style
\emph{direction arithmetic} (Section~\ref{sec:directional}) toward
either a supervised direction or an emergent mode pole.

\subsection{Pairings}\label{sec:pairings}

In order to test how each embedding answers the simplest culinary
question -- what pairs with X -- we compute the top-5 cosine-nearest
neighbours plus the closest emergent mode (cosine to mode pole) for
twelve canonical probe seeds, with FlavorGraph as an external foil.

We found that the three Epicure models return culinarily coherent
peers at consistent granularity while FlavorGraph returns long
preparation-level strings that fragment the co-occurrence signal
across scattered food (and non-food) vocabulary (Table~\ref{tab:neighbors}).

\begin{landscape}
{\small
\setlength{\tabcolsep}{4pt}
\begin{longtable}{@{}lp{5.4cm}p{5.4cm}p{5.4cm}p{5.4cm}@{}}
\caption{Top-5 nearest neighbours by cosine similarity.  The FlavorGraph column is retained as an illustrative foil -- its 6{,}653-ingredient vocabulary contains brand names and preparation modifiers (e.g.~``sourdough roll'', ``kraft shredded triple cheddar cheese with a touch of philadelphia'') that our 1{,}790-canonical vocabulary collapses, so FG's neighbours are drawn from a structurally different name space.}
\label{tab:neighbors} \\
\toprule
\textbf{Ingredient} & \textbf{Epicure-Cooc top-5} & \textbf{Epicure-Core top-5} & \textbf{Epicure-Chem top-5} & \textbf{FlavorGraph top-5} \\
\midrule
\endfirsthead
\multicolumn{5}{@{}l}{\small\itshape Table~\ref{tab:neighbors} continued} \\[2pt]
\toprule
\textbf{Ingredient} & \textbf{Epicure-Cooc top-5} & \textbf{Epicure-Core top-5} & \textbf{Epicure-Chem top-5} & \textbf{FlavorGraph top-5} \\
\midrule
\endhead
\midrule
\multicolumn{5}{r@{}}{\small\itshape continued on next page} \\
\endfoot
\bottomrule
\endlastfoot
chicken & garlic (0.39), onion (0.37), black pepper (0.36), turkey (0.35), carrot (0.34) & pork (0.58), beef (0.57), chicken broth (0.55), peanut (0.52), cream of chicken soup (0.52) & beef (0.41), pork (0.34), cream of chicken soup (0.31), buffalo wing sauce (0.29), peanut (0.28) & sourdough roll (0.83), macaroni shells and cheese (0.83), kraft shredded triple cheddar cheese with a touch of philadelphia (0.82), heads of garlic (0.81), montreal chicken seasoning (0.79) \\
salmon & wasabi (0.35), avocado (0.32), ponzu (0.32), balsamic vinegar (0.32), scallop (0.32) & trout (0.65), ham (0.63), mirin (0.62), cod (0.62), fish roe (0.61) & ham (0.50), trout (0.36), tuna (0.34), cod (0.33), miso (0.31) & lemon pepper seasoning (0.91), wasabi paste (0.85), dried dill weed (0.84), liquid smoke flavoring (0.74), dried dill (0.73) \\
tomato & onion (0.48), parsley (0.47), garlic (0.46), olive oil (0.46), bell pepper (0.44) & bell pepper (0.61), red pepper (0.60), olive oil (0.59), onion (0.59), red onion (0.58) & bean (0.29), bell pepper (0.29), red pepper (0.29), tortilla (0.28), onion (0.28) & crisp salad green (0.90), dry chili pepper (0.89), light balsamic vinaigrette salad dressing (0.89), western salad dressing (0.89), tex mex cheese (0.89) \\
basil & parsley (0.45), olive oil (0.44), parmesan cheese (0.44), black pepper (0.42), white wine (0.39) & oregano (0.71), tarragon (0.68), rosemary (0.67), olive oil (0.67), pasta (0.67) & tarragon (0.46), oregano (0.46), rosemary (0.44), pasta (0.40), fennel (0.38) & light mozzarella cheese (0.78), minced garlic clove (0.75), italian plum tomato (0.74), italian cut green bean (0.73), fat free parmesan cheese (0.72) \\
chocolate & cocoa powder (0.55), vanilla (0.49), milk chocolate (0.48), almond (0.48), white chocolate (0.47) & cocoa powder (0.85), toffee (0.79), frosting (0.77), fudge (0.74), cream (0.73) & toffee (0.69), cocoa powder (0.68), frosting (0.67), cream (0.65), marshmallow (0.64) & pen (0.84), mint extract (0.82), espresso powder (0.81), tea biscuit (0.80), dark cocoa (0.71) \\
rice & carrot (0.34), okra (0.33), vegetable oil (0.33), pea (0.31), dashi (0.31) & shiitake mushroom (0.60), nori (0.59), bok choy (0.58), brown rice (0.58), sesame seed (0.57) & caviar (0.31), nori (0.30), wheat (0.30), millet (0.30), puzi leaf (0.29) & frozen peas and carrot (0.93), boneless skinless chicken (0.82), chicken giblet (0.77), dark soya sauce (0.76), chicken part (0.71) \\
butter & cream (0.46), egg yolk (0.43), flour (0.41), milk (0.41), almond (0.38) & milk (0.60), egg yolk (0.57), egg (0.56), parmesan cheese (0.50), white chocolate (0.49) & milk (0.48), flour (0.35), egg (0.33), chocolate (0.33), parmesan cheese (0.31) & almond macaroon (0.86), snow crab leg (0.85), ginger in syrup (0.85), caramel square (0.85), yellow cling peach (0.84) \\
soy sauce & sesame oil (0.54), shiitake mushroom (0.47), ginger (0.47), oyster sauce (0.46), light soy sauce (0.45) & sesame oil (0.75), light soy sauce (0.71), oyster sauce (0.70), shaoxing wine (0.70), doubanjiang (0.69) & scallion (0.45), enoki mushroom (0.35), bamboo shoot (0.32), sake (0.32), bok choy (0.32) & red rice (0.92), ginkgo nut (0.72), rice cake (0.58), chinese mustard (0.54), guava (0.52) \\
lemon & orange (0.42), mint (0.41), rosemary (0.40), cherry (0.40), mustard (0.38) & lime (0.65), orange (0.62), thyme (0.58), fennel (0.57), olive oil (0.57) & lime (0.49), clementine (0.44), orange (0.43), pomegranate (0.38), parsley (0.37) & white fleshed fish (0.88), lemon sherbet (0.88), rockfish fillet (0.88), capers in brine (0.87), baby chicken (0.87) \\
cumin & garlic (0.44), coriander (0.43), bay leaf (0.43), chili powder (0.42), black pepper (0.39) & chili powder (0.74), coriander (0.70), turmeric (0.70), chickpea (0.67), green chili (0.67) & turmeric (0.45), ajwain (0.41), black pepper (0.40), parsley (0.40), chili powder (0.40) & hatch chile (0.87), mexican tomato sauce (0.81), dark mexican beer (0.79), mexican style diced tomato (0.79), green taco sauce (0.78) \\
shrimp & chili pepper (0.36), sesame oil (0.36), clam (0.36), cabbage (0.35), squid (0.34) & squid (0.67), clam (0.67), oyster (0.62), crab stick (0.62), crab (0.60) & clam (0.43), oyster (0.38), squid (0.37), krill (0.36), crab (0.31) & firm white fish (0.76), cooked grit (0.76), quick cooking grit (0.73), grit (0.73), old bay seasoning (0.64) \\
lentil & turmeric (0.35), cayenne pepper (0.32), mustard seed (0.31), pomegranate molasses (0.31), sambar powder (0.31) & chickpea (0.79), vegetable stock (0.73), butternut squash (0.69), olive oil (0.68), zucchini (0.68) & chickpea (0.61), butternut squash (0.54), quinoa (0.49), zucchini (0.49), kale (0.49) & fresh bay leaf (0.75), dried garbanzo bean (0.75), whole bay leaf (0.72), dried navy bean (0.64), long grain brown rice (0.62) \\
\end{longtable}
}
\end{landscape}

We also found that the three Epicure models retrieve different
\emph{kinds} of neighbour for the same seed.  For chicken, Cooc's
top hit is \emph{garlic} (recipe companion) and its full top-5 is
mostly aromatic vegetables (\emph{garlic, onion, black\_pepper,
turkey, carrot}); Core's top hit is \emph{pork} (chemistry peer);
Chem's is \emph{beef}, and its top-5 sits in the chicken-cooking
neighbourhood -- two protein peers plus three canonical chicken
accompaniments (\emph{beef, pork, cream\_of\_chicken\_soup,
buffalo\_wing\_sauce, peanut}).  For basil, Cooc retrieves
\emph{parsley} (co-occurrence peer); Core \emph{oregano} and Chem
\emph{tarragon} both sit in the Italian-herb chemistry cluster and
share four of five top-5 peers (\emph{oregano, tarragon, rosemary,
pasta}), while Cooc's top-5 reaches for basil's pasta-pantry context
(\emph{olive\_oil, parmesan\_cheese, black\_pepper, white\_wine}).
This means the three siblings expose the two paths a chef might take
when reaching for a replacement: ``what else do I cook with this''
(Cooc) versus ``what shares its flavour profile'' (Core and Chem).

\paragraph{Mode-membership pairings.}
Table~\ref{tab:pairings-catalogue} extends the simple top-$K$
neighbour view to mode-membership lookup: for a probe seed, the
closest emergent mode in each model (cosine to mode pole) together
with the other top members of that mode.  This separates ``where in
the atlas does the seed live'' from ``what's nearest to the seed''
-- a chef-facing tool typically wants both.

\begin{landscape}
{\footnotesize
\setlength{\tabcolsep}{3pt}
\renewcommand{\arraystretch}{0.85}
\begin{longtable}{l>{\raggedright\arraybackslash}p{5.2cm}>{\raggedright\arraybackslash}p{5.2cm}>{\raggedright\arraybackslash}p{5.2cm}}
\caption{Emergent pairings catalogue: for each probe seed, the closest emergent ($F_*$) mode the seed sits in for each model, with the other top members of that mode.  The sweet-confection neighbourhood for chocolate is shared across all three models, while cuisine framing for savoury and umami seeds (tomato, miso, preserved lemon) shifts across the chemistry/co-occurrence axis.}
\label{tab:pairings-catalogue} \\
\toprule
\textbf{Seed} & \textbf{Cooc} & \textbf{Core} & \textbf{Chem} \\
\midrule
\endfirsthead
\multicolumn{4}{@{}l}{\footnotesize\itshape Table~\ref{tab:pairings-catalogue} continued} \\[2pt]
\toprule
\textbf{Seed} & \textbf{Cooc} & \textbf{Core} & \textbf{Chem} \\
\midrule
\endhead
\midrule
\multicolumn{4}{r@{}}{\footnotesize\itshape continued on next page} \\
\endfoot
\bottomrule
\endlastfoot
chocolate & \texttt{F\_4/M3}\newline \textbf{chocolate and coffee confections} (cos=+0.69)\newline ganache\newline coffee liqueur\newline cocoa powder & \texttt{F\_15/M0}\newline \textbf{American confectionery and sweet treats} (cos=+0.86)\newline graham cracker\newline toffee\newline fudge & \texttt{F\_5/M5}\newline \textbf{confectionery and dessert components} (cos=+0.84)\newline gelatin\newline cream\newline meringue \\
\cmidrule(l){2-4}
tomato & \texttt{F\_8/M0}\newline \textbf{Everyday Western savory vegetables} (cos=+0.68)\newline black pepper\newline onion\newline lettuce & \texttt{F\_0/M0}\newline \textbf{Mediterranean savory cooking staples} (cos=+0.67)\newline vegetable stock\newline garlic\newline zucchini & \texttt{F\_5/M1}\newline \textbf{Hearty bean and sausage stew ingredients} (cos=+0.32)\newline red onion\newline cannellini bean\newline bell pepper \\
\cmidrule(l){2-4}
miso & \texttt{F\_10/M1}\newline \textbf{Japanese hot pot ingredients} (cos=+0.44)\newline enoki mushroom\newline garland chrysanthemum\newline shiitake mushroom & \texttt{F\_8/M5}\newline \textbf{Japanese vegetables and umami seasonings} (cos=+0.81)\newline rice vinegar\newline enoki mushroom\newline oyster mushroom & \texttt{F\_13/M4}\newline \textbf{Savory protein-rich seafood and cheese} (cos=+0.55)\newline cheese\newline cod\newline octopus \\
\cmidrule(l){2-4}
lentil & \texttt{F\_2/M2}\newline \textbf{Mediterranean pantry seeds and aromatics} (cos=+0.40)\newline pomegranate\newline salt\newline pumpkin seed & \texttt{F\_12/M3}\newline \textbf{Mediterranean whole-food vegetables and grains} (cos=+0.79)\newline balsamic vinegar\newline chicken broth\newline fennel seed & \texttt{F\_7/M4}\newline \textbf{Mediterranean garden vegetables and grains} (cos=+0.58)\newline kefir\newline beet\newline semolina \\
\cmidrule(l){2-4}
preserved lemon & \texttt{F\_2/M2}\newline \textbf{Mediterranean pantry seeds and aromatics} (cos=+0.29)\newline pomegranate\newline salt\newline pumpkin seed & \texttt{F\_14/M0}\newline \textbf{Middle Eastern spice blends and staples} (cos=+0.58)\newline berbere\newline falafel\newline baharat & \texttt{F\_14/M4}\newline \textbf{Mediterranean savory pantry staples} (cos=+0.29)\newline olive oil\newline cayenne pepper\newline balsamic vinegar \\
\cmidrule(l){2-4}
coffee & \texttt{F\_6/M2}\newline \textbf{sweet cocktail and confection ingredients} (cos=+0.68)\newline rum\newline white chocolate\newline ganache & \texttt{F\_19/M3}\newline \textbf{Sweet dessert liqueurs and confections} (cos=+0.58)\newline coffee liqueur\newline chocolate liqueur\newline maraschino cherry & \texttt{F\_13/M3}\newline \textbf{Sweet liqueurs and cocktail ingredients} (cos=+0.45)\newline coffee liqueur\newline liqueur\newline hazelnut liqueur \\
\end{longtable}
}
\end{landscape}

Nearest neighbours expose where the seed already sits; steering the
seed in a chosen direction requires an explicit operator, which the
next subsection introduces.

\subsection{Direction arithmetic}\label{sec:directional}

In order to test whether a seed can be steered along a culinary axis
-- and how cleanly that motion respects either supervised labels or
unsupervised mode geometry -- we apply SLERP rotation of the seed
toward a unit direction by angle $\theta$ on the unit sphere.  At
$0^{\circ}$ the rotated query is the unmodified seed; at
$60^{\circ}$ its cosine similarity to the seed has dropped to
$0.5$ and the target's neighbourhood dominates.  Two direction
families are available: \emph{supervised} pole vectors built from
labelled tags (cuisine macro-regions, food groups, NOVA processing
class), and the \emph{emergent} factor-mode poles from
Section~\ref{sec:emergent}.

\paragraph{SLERP toward supervised directions.}
Table~\ref{tab:direction-arithmetic} reports four hero rotations
toward supervised pole vectors at $30^{\circ}$ and
$60^{\circ}$.  We found that the destinations are label-aligned
in every model: rice rotated toward the South-Asian direction at
$30^{\circ}$ retrieves \emph{curry leaf, masoor dal, urad dal,
chana dal, fenugreek seed} in Cooc; corn rotated toward Latin
American at $30^{\circ}$ retrieves \emph{salsa verde, tomatillo,
queso fresco, fajita seasoning, corn tortilla}.  Multi-constraint
queries -- chicken rotated toward processed $+$ Western Atlantic at
$60^{\circ}$ -- converge on mid-century American home-cooking
staples (\emph{swiss cheese, steak sauce, turkey, sour cream, ranch
dressing} in Cooc; \emph{cheddar cheese, cream of chicken soup,
crescent roll, alfredo sauce, ranch dressing} in Core; \emph{colby
cheese, buffalo wing sauce, ranch dressing, cream of chicken soup,
alfredo sauce} in Chem).  This means supervised SLERP is a
predictable, label-aligned steering operator across all three
siblings.

\begin{landscape}
{\footnotesize
\setlength{\tabcolsep}{3pt}
\renewcommand{\arraystretch}{0.85}
\begin{longtable}{@{}>{\raggedright\arraybackslash}p{4cm}l>{\raggedright\arraybackslash}p{5cm}>{\raggedright\arraybackslash}p{5cm}>{\raggedright\arraybackslash}p{5cm}@{}}
\caption{Direction arithmetic (SLERP), hero cases for the main paper.  Seed is rotated toward the learned direction on the unit sphere by the specified angle; top-5 nearest neighbours of the rotated query vector are reported. $0^{\circ}$ is the unmodified seed, $60^{\circ}$ is a full rotation (cosine similarity to seed $=$ 0.50). See \texttt{direction\_arithmetic\_full} for all 48 test cases.}
\label{tab:direction-arithmetic} \\
\toprule
\textbf{Test case} & \textbf{Rotation} & \textbf{Epicure-Cooc top-5} & \textbf{Epicure-Core top-5} & \textbf{Epicure-Chem top-5} \\
\midrule
\endfirsthead
\multicolumn{5}{@{}l}{\footnotesize\itshape Table~\ref{tab:direction-arithmetic} continued} \\[2pt]
\toprule
\textbf{Test case} & \textbf{Rotation} & \textbf{Epicure-Cooc top-5} & \textbf{Epicure-Core top-5} & \textbf{Epicure-Chem top-5} \\
\midrule
\endhead
\midrule
\multicolumn{5}{r@{}}{\footnotesize\itshape continued on next page} \\
\endfoot
\bottomrule
\endlastfoot
rice + South\_Asian & $0^{\circ}$ & carrot\newline okra\newline vegetable oil\newline pea\newline dashi & shiitake mushroom\newline nori\newline bok choy\newline sesame seed\newline pea & caviar\newline nori\newline wheat\newline millet\newline puzi leaf \\
\cmidrule(l){2-5}
 & $30^{\circ}$ & curry leaf\newline masoor dal\newline urad dal\newline chana dal\newline fenugreek seed & chana dal\newline fenugreek leaf\newline urad dal\newline toor dal\newline horse gram & chana dal\newline toor dal\newline fenugreek seed\newline kashmiri chili\newline sambar powder \\
\cmidrule(l){2-5}
 & $60^{\circ}$ & curry leaf\newline chana dal\newline urad dal\newline masoor dal\newline horse gram & chana dal\newline toor dal\newline kashmiri chili\newline urad dal\newline horse gram & chana dal\newline kashmiri chili\newline toor dal\newline fenugreek seed\newline amchur \\
\midrule
\multicolumn{5}{@{}l}{\textit{}} \\
\midrule
corn + Latin\_American & $0^{\circ}$ & pea\newline scallion\newline shrimp\newline red onion\newline rice & bell pepper\newline potato\newline carrot\newline pinto bean\newline cheddar cheese & tortilla\newline guascas\newline red pepper\newline oregano\newline epazote \\
\cmidrule(l){2-5}
 & $30^{\circ}$ & salsa verde\newline tomatillo\newline queso fresco\newline fajita seasoning\newline corn tortilla & tomatillo\newline corn tortilla\newline epazote\newline salsa\newline nopal & corn tortilla\newline queso fresco\newline epazote\newline salsa\newline enchilada sauce \\
\cmidrule(l){2-5}
 & $60^{\circ}$ & tomatillo\newline queso fresco\newline corn tortilla\newline salsa verde\newline refried beans & tomatillo\newline corn tortilla\newline nopal\newline queso fresco\newline epazote & corn tortilla\newline queso fresco\newline salsa\newline guajillo chile\newline poblano pepper \\
\midrule
\multicolumn{5}{@{}l}{\textit{}} \\
\midrule
chicken + processed + Western\_Atlantic & $0^{\circ}$ & garlic\newline onion\newline black pepper\newline turkey\newline carrot & pork\newline beef\newline chicken broth\newline peanut\newline cream of chicken soup & beef\newline pork\newline cream of chicken soup\newline buffalo wing sauce\newline peanut \\
\cmidrule(l){2-5}
 & $30^{\circ}$ & turkey\newline swiss cheese\newline onion\newline steak sauce\newline sour cream & cheddar cheese\newline cream of chicken soup\newline alfredo sauce\newline beef\newline turkey & cream of chicken soup\newline buffalo wing sauce\newline beef\newline ranch dressing\newline cheddar cheese \\
\cmidrule(l){2-5}
 & $60^{\circ}$ & swiss cheese\newline steak sauce\newline turkey\newline sour cream\newline ranch dressing & cheddar cheese\newline cream of chicken soup\newline crescent roll\newline alfredo sauce\newline ranch dressing & colby cheese\newline buffalo wing sauce\newline ranch dressing\newline cream of chicken soup\newline alfredo sauce \\
\midrule
\multicolumn{5}{@{}l}{\textit{}} \\
\midrule
bread + high protein & $0^{\circ}$ & olive oil\newline parsley\newline paprika\newline thyme\newline anchovy & pasta\newline artichoke\newline olive oil\newline parsley\newline red onion & pasta\newline caper\newline artichoke\newline anchovy\newline provolone cheese \\
\cmidrule(l){2-5}
 & $30^{\circ}$ & olive oil\newline parsley\newline black pepper\newline pasta\newline paprika & olive oil\newline pasta\newline artichoke\newline red onion\newline anchovy & pasta\newline artichoke\newline provolone cheese\newline anchovy\newline caper \\
\cmidrule(l){2-5}
 & $60^{\circ}$ & black pepper\newline pasta\newline tomato\newline parmesan cheese\newline pepperoni & anchovy\newline pita bread\newline olive oil\newline tuna\newline sausage & provolone cheese\newline monkfish\newline anchovy\newline hot dog\newline pita bread \\
\end{longtable}
}
\end{landscape}

\paragraph{SLERP toward emergent mode poles.}
The same SLERP operator works on emergent targets.
Table~\ref{tab:rotate-hero} reports three rotations from various
seeds toward an \emph{intent} -- a target concept resolved per
model to its best-matching mode by label keyword.  The target
$(F_X, M_Y)$ coordinate differs across models because ICA
orientations are model-specific; the cells show the actual
coordinate used and the top-5 hits.  We found that the destinations
differ across models in ways that mirror their geometry.
\emph{chocolate} rotated toward sweet baking lands on a
baking-and-confection cluster in all three models, though the
cultural framing differs: Cooc and Core both reach a Western
sweet-baking neighbourhood (\emph{cocoa\_powder, vanilla, coffee}
for Cooc; \emph{baking\_powder, chia\_seed, whole\_wheat\_flour}
for Core), while Chem lands on an East-Asian dessert mode anchored
by \emph{red\_bean\_paste, matcha\_powder, purple\_sweet\_potato}.
\emph{chicken} rotated toward Southeast-Asian aromatics traces the
same chemistry/co-occurrence split: Cooc picks an Indonesian
spice-paste mode (\emph{candlenut, kencur, garam\_masala}), Core a
broader East/Southeast-Asian pantry mode (\emph{rice\_noodle,
bean\_sprout, fish\_ball}), and Chem a Southeast-Asian chili-spice
mode (\emph{chili\_pepper, sichuan\_peppercorn,
birds\_eye\_chili}).  \emph{tomato} rotated toward a Mediterranean
savoury pantry retrieves model-specific regional cuts of the same
concept: a savoury whole-food Mediterranean staples mode in Cooc
(\emph{turkey, butternut\_squash, kale}), an Eastern Mediterranean
cheese-and-flatbread mode in Core (\emph{tulum\_cheese,
kasseri\_cheese, yufka}), and a Caucasian--Mediterranean pantry
mode in Chem (\emph{sulguni\_cheese, sun\_dried\_tomato, adjika}).
This means emergent SLERP exposes each model's training bias --
Cooc reaching recipe-context neighbours, Chem reaching
chemistry-clustered ones -- as a navigable knob rather than hiding
it.

{\footnotesize
\setlength{\tabcolsep}{3pt}
\renewcommand{\arraystretch}{0.85}
\begin{longtable}{l>{\raggedright\arraybackslash}p{2.4cm}>{\raggedright\arraybackslash}p{4.0cm}>{\raggedright\arraybackslash}p{4.0cm}>{\raggedright\arraybackslash}p{4.0cm}}
\caption{Hero \emph{mode\_rotate} examples by SHARED INTENT, not shared coordinate.  Each row pins a target concept (e.g.\ \emph{sweet baking / confection}) and resolves it to the model-specific mode whose label best matches; the table row then rotates the seed toward THAT model's mode, showing the actual (factor, mode\_id) coordinate used and the resulting top-5 hits.  Rows therefore compare same-concept rotations across the three Epicure models, not same-coordinate rotations.  The angle is $60^{\circ}$ throughout.  See Table~\ref{tab:rotate-angle-sweep} for an angle sweep showing how the seed identity fades as the angle grows.}
\label{tab:rotate-hero} \\
\toprule
\textbf{Seed} & \textbf{Intent} & \textbf{Cooc} & \textbf{Core} & \textbf{Chem} \\
\midrule
\endfirsthead
\multicolumn{5}{@{}l}{\footnotesize\itshape Table~\ref{tab:rotate-hero} continued} \\[2pt]
\toprule
\textbf{Seed} & \textbf{Intent} & \textbf{Cooc} & \textbf{Core} & \textbf{Chem} \\
\midrule
\endhead
\midrule
\multicolumn{5}{r@{}}{\footnotesize\itshape continued on next page} \\
\endfoot
\bottomrule
\endlastfoot
chocolate & Sweet baking / confection & \texttt{F\_7/M1}\newline \textbf{Sweet baking and plant-based pantry staples}\newline cocoa powder (+0.66)\newline vanilla (+0.65)\newline coffee (+0.65)\newline hazelnut (+0.64)\newline cacao (+0.63) & \texttt{F\_5/M2}\newline \textbf{Sweet baking and confectionery ingredients}\newline baking powder (+0.81)\newline chia seed (+0.81)\newline whole wheat flour (+0.78)\newline baking soda (+0.77)\newline soy protein isolate (+0.76) & \texttt{F\_12/M1}\newline \textbf{East Asian confectionery and sweet baking ingredients}\newline red bean paste (+0.72)\newline matcha powder (+0.71)\newline purple sweet potato (+0.69)\newline mochi (+0.66)\newline mung bean paste (+0.63) \\
\cmidrule(l){2-5}
chicken & Southeast-Asian aromatics & \texttt{F\_0/M0}\newline \textbf{Indonesian spice paste aromatics}\newline candlenut (+0.66)\newline kencur (+0.65)\newline garam masala (+0.63)\newline palm sugar (+0.61)\newline sweet soy sauce (+0.60) & \texttt{F\_13/M0}\newline \textbf{East and Southeast Asian pantry staples}\newline rice noodle (+0.89)\newline bean sprout (+0.88)\newline fish ball (+0.88)\newline udon noodle (+0.87)\newline sesame oil (+0.87) & \texttt{cf\_minty/M4}\newline \textbf{Southeast Asian aromatic chili spices}\newline chili pepper (+0.73)\newline sichuan peppercorn (+0.72)\newline birds eye chili (+0.67)\newline shaoxing wine (+0.64)\newline dark soy sauce (+0.63) \\
\cmidrule(l){2-5}
tomato & Mediterranean savoury pantry & \texttt{F\_7/M2}\newline \textbf{Savory whole-food Mediterranean pantry staples}\newline turkey (+0.63)\newline butternut squash (+0.62)\newline kale (+0.62)\newline vegetable stock (+0.61)\newline portobello mushroom (+0.61) & \texttt{F\_1/M1}\newline \textbf{Eastern Mediterranean pantry staples}\newline tulum cheese (+0.87)\newline kasseri cheese (+0.85)\newline yufka (+0.83)\newline ezine cheese (+0.82)\newline kashkaval cheese (+0.81) & \texttt{F\_7/M1}\newline \textbf{Caucasian and Eastern Mediterranean pantry staples}\newline sulguni cheese (+0.66)\newline sun dried tomato (+0.65)\newline adjika (+0.63)\newline khmeli suneli (+0.63)\newline bryndza (+0.63) \\
\end{longtable}
}

\paragraph{The angle is a continuous knob.}
Table~\ref{tab:rotate-angle-sweep} demonstrates how the rotated
query transitions from seed-dominated to target-dominated as the
angle grows.  Two seeds (chicken and beef) rotate toward a single
canonical chef intent -- the \emph{Mexican / Tex-Mex pantry} mode
(chicken fajitas / beef barbacoa territory) -- at three angles
($0^{\circ}$, $30^{\circ}$, $60^{\circ}$) in each Epicure model.  We found that
at $0^{\circ}$ the rotated query is the unmodified seed and the
top-5 is the seed's own nearest neighbourhood (Cooc beef returns
\emph{onion, pork, black\_pepper, garlic, potato}; Core chicken
returns \emph{pork, beef, chicken\_broth, peanut,
cream\_of\_chicken\_soup}); by $30^{\circ}$ Tex-Mex
intermediates dominate (Cooc beef: \emph{corn\_tortilla,
monterey\_jack\_cheese, onion, pinto\_bean, salsa}; Core chicken:
\emph{monterey\_jack\_cheese, flour\_tortilla, corn\_tortilla,
salsa\_verde, enchilada\_sauce}); at $60^{\circ}$ both seeds
collapse onto a nearly identical Mexican-specialty neighbourhood
-- in Core both retrieve the same Tex-Mex top-5
(\emph{corn\_tortilla, salsa, monterey\_jack\_cheese,
flour\_tortilla, tortilla}); in Cooc both share
\emph{corn\_tortilla, monterey\_jack\_cheese, salsa\_verde,
salsa, poblano\_pepper}; in Chem both share \emph{poblano\_pepper,
salsa, cotija\_cheese, corn\_tortilla, monterey\_jack\_cheese}.
The $60^{\circ}$ destinations are specialty Mexican
ingredients (\emph{cotija\_cheese, ancho\_chile, poblano\_pepper,
salsa\_verde}) the seeds themselves do not retrieve directly; the
rotation surfaces them from a generic meat seed.  This means the
angle is a continuous dial between seed and target, and chef-facing
tools should expose it so a user can stay close to the seed when
refining or travel further when exploring.

{\footnotesize
\setlength{\tabcolsep}{3pt}
\renewcommand{\arraystretch}{0.85}
\begin{longtable}{lc>{\raggedright\arraybackslash}p{4.0cm}>{\raggedright\arraybackslash}p{4.0cm}>{\raggedright\arraybackslash}p{4.0cm}}
\caption{Angle sweep on \emph{mode\_rotate}: chicken and beef rotated toward the \emph{Mexican / Tex-Mex pantry} intent in each Epicure model at three angles (0\textdegree, 30\textdegree, 60\textdegree).  At 0\textdegree the rotated query is the unmodified seed and the top-5 is the seed's own nearest neighbourhood; as the angle grows the query moves toward the target pole and at 60\textdegree the two seeds collapse onto a nearly identical target neighbourhood, demonstrating that the angle is a continuous knob between \emph{stay near the seed} and \emph{go to the target}.  All other rows in Table~\ref{tab:rotate-hero} use 60\textdegree.}
\label{tab:rotate-angle-sweep} \\
\toprule
\textbf{Seed} & \textbf{Angle} & \textbf{Cooc} & \textbf{Core} & \textbf{Chem} \\
\midrule
\endfirsthead
\multicolumn{5}{@{}l}{\footnotesize\itshape Table~\ref{tab:rotate-angle-sweep} continued} \\[2pt]
\toprule
\textbf{Seed} & \textbf{Angle} & \textbf{Cooc} & \textbf{Core} & \textbf{Chem} \\
\midrule
\endhead
\midrule
\multicolumn{5}{r@{}}{\footnotesize\itshape continued on next page} \\
\endfoot
\bottomrule
\endlastfoot
\midrule
\multicolumn{5}{@{}l}{\textit{Seed: \emph{chicken}}} \\
\midrule
chicken & 0\textdegree & garlic (+0.39)\newline onion (+0.37)\newline black pepper (+0.36)\newline turkey (+0.35)\newline carrot (+0.34) & pork (+0.58)\newline beef (+0.57)\newline chicken broth (+0.55)\newline peanut (+0.52)\newline cream of chicken soup (+0.52) & beef (+0.41)\newline pork (+0.34)\newline cream of chicken soup (+0.31)\newline buffalo wing sauce (+0.29)\newline peanut (+0.28) \\
\cmidrule(l){2-5}
chicken & 30\textdegree & corn tortilla (+0.54)\newline salsa (+0.51)\newline monterey jack cheese (+0.50)\newline fajita seasoning (+0.50)\newline salsa verde (+0.50) & monterey jack cheese (+0.80)\newline flour tortilla (+0.79)\newline corn tortilla (+0.79)\newline salsa verde (+0.77)\newline enchilada sauce (+0.77) & enchilada sauce (+0.51)\newline salsa (+0.51)\newline flour tortilla (+0.51)\newline corn tortilla (+0.51)\newline salsa verde (+0.49) \\
\cmidrule(l){2-5}
chicken & 60\textdegree & corn tortilla (+0.67)\newline monterey jack cheese (+0.66)\newline salsa verde (+0.65)\newline salsa (+0.63)\newline poblano pepper (+0.63) & corn tortilla (+0.91)\newline salsa (+0.91)\newline monterey jack cheese (+0.89)\newline flour tortilla (+0.89)\newline tortilla (+0.89) & corn tortilla (+0.69)\newline poblano pepper (+0.68)\newline salsa (+0.68)\newline cotija cheese (+0.67)\newline ancho chile (+0.67) \\
\midrule
\multicolumn{5}{@{}l}{\textit{Seed: \emph{beef}}} \\
\midrule
beef & 0\textdegree & onion (+0.41)\newline pork (+0.37)\newline black pepper (+0.35)\newline garlic (+0.35)\newline potato (+0.32) & chicken (+0.57)\newline taco seasoning (+0.56)\newline bell pepper (+0.55)\newline enchilada sauce (+0.55)\newline mushroom (+0.54) & chicken (+0.41)\newline taco sauce (+0.34)\newline taco seasoning (+0.33)\newline enchilada sauce (+0.30)\newline mushroom (+0.30) \\
\cmidrule(l){2-5}
beef & 30\textdegree & corn tortilla (+0.52)\newline monterey jack cheese (+0.52)\newline onion (+0.48)\newline pinto bean (+0.48)\newline salsa (+0.47) & salsa (+0.81)\newline flour tortilla (+0.81)\newline monterey jack cheese (+0.81)\newline enchilada sauce (+0.80)\newline corn tortilla (+0.79) & enchilada sauce (+0.55)\newline salsa (+0.53)\newline pinto bean (+0.52)\newline monterey jack cheese (+0.51)\newline flour tortilla (+0.49) \\
\cmidrule(l){2-5}
beef & 60\textdegree & corn tortilla (+0.67)\newline monterey jack cheese (+0.67)\newline salsa verde (+0.63)\newline poblano pepper (+0.62)\newline salsa (+0.62) & corn tortilla (+0.91)\newline salsa (+0.90)\newline monterey jack cheese (+0.89)\newline flour tortilla (+0.89)\newline tortilla (+0.89) & poblano pepper (+0.70)\newline salsa (+0.68)\newline cotija cheese (+0.68)\newline corn tortilla (+0.67)\newline monterey jack cheese (+0.67) \\
\end{longtable}
}

Supervised SLERP gives label-aligned steering; emergent SLERP gives
steering without curated labels; the angle is a continuous dial
between seed and target.  Section~\ref{sec:discussion} considers
what corpus and operator extensions these primitives suggest.

\section{Discussion}\label{sec:discussion}

\subsection{What the controlled comparison shows}
\label{sec:disc-controlled-comparison}

Cooc, Core, and Chem share architecture, hyperparameters, vocabulary,
graph node set, and the entire $203{,}508$-edge co-occurrence
backbone (Section~\ref{sec:methods}); they differ only in which
typed walks the skip-gram objective sees and at what rate.  Two
findings follow from holding everything else fixed.  First, the
Cooc~$<$~Core~$<$~Chem ordering of supervised direction quality
(Section~\ref{sec:direction-quality}) holds on every probe stratum
we test, including the five basic-taste, eight USDA-macronutrient,
and eight cuisine-macro-region probes that the compound-feature
schema never sees.  Chemistry-mediated walks therefore act as a
structural prior whose reach extends beyond the labels they
directly encode: routing context through shared aroma compounds
makes a broader family of culinary concepts linearly recoverable
than the schema names, and Mikolov-style linear
directions~\citep{mikolov2013distributed} are the mechanism by which
that prior becomes geometry.  Second, Core's concentrated geometry
(participation ratio $94.2$ against Cooc's $173.6$ and Chem's
$183.1$; Section~\ref{sec:isotropy-foodgroup}) is a deliberate
consequence of the $10\times$ I--I walk injection, not a
corpus-induced collapse of the kind \citet{mu2018allbutthetop}
address.  It coincides with stronger linear probes than either
isotropic sibling and with the tightest emergent modes of the
three (Section~\ref{sec:emergent}), so the concentration is a
design lever rather than a defect to rescue.

\subsection{From recommendation to navigation}
\label{sec:disc-navigation}

The chemistry-vs-recipe-context axis surfaces twice in the operator
output of Section~\ref{sec:transformations}: at the nearest-neighbour
level the same seed returns a recipe companion under Cooc and a
flavour-profile peer under Chem, and at the SLERP-destination level
the same seed and target angle land on culturally different framings
of the target concept depending on the sibling.  The user-facing
primitives therefore decompose into three independent choices, all
expressed on the same 300-D embedding: which sibling to query (which
question is being asked, co-occurrence companion or flavour-profile
peer), which direction to rotate toward (a supervised pole vector or
an emergent factor-mode pole), and how far to travel (the SLERP
angle).  Closest-mode lookup
(Section~\ref{sec:pairings}) gives users the named-cluster query a
knowledge graph like FoodKG~\citep{haussmann2019foodkg} would offer
(\emph{which named region is this ingredient in?}) without
sacrificing the continuous-geometry query that an embedding like
FlavorGraph~\citep{park2021flavorgraph} is designed for; the two
affordances live on the same 300-D model rather than in separate
systems.  The methodological move that makes this possible --
treating the walk schema as a named axis rather than an architectural
constant -- applies to any future fusion
of chemistry, nutrient, sensory, image, or recipe-text signals.

\subsection{Limitations}
\label{sec:disc-limitations}

\paragraph{Corpus imbalance.}
The 4.14M-recipe corpus is roughly half East Asian and a tenth
Mediterranean, with single-digit shares for South Asian, Eastern
European, and Latin American cuisines (Section~\ref{sec:data-sources}).
The held-out-cuisine $d$ confidence intervals
(Figure~\ref{fig:cuisine-d}) widen accordingly in the smaller
regions; the cross-region \emph{ranking} of the three siblings is
nevertheless stable, so the imbalance limits resolution within a
region more than it threatens the synthesis above.

\paragraph{Hub coverage.}
$525$ of $1{,}790$ canonical ingredients anchor against FlavorDB
under our entity-unique matching policy ($523$ retain active I--C
edges after the \texttt{min\_compound\_degree=2} filter); the
remaining $1{,}267$ non-hubs participate in both Core and Chem,
but they reach compound context only indirectly, through the
via-compound metapath \texttt{N--H--C[x]--H--N}
(Section~\ref{sec:models}) that bridges two non-hubs through a
hub--compound--hub spine and contributes the bulk of Chem's
skip-gram pair budget.  Their chemistry signal is therefore one
walk-hop further removed from the compound vertex than that of the
$523$ hubs; broader compound coverage (FooDB, USDA Food Patterns
Equivalents) would promote more non-hubs to hub status and shorten
that chain.

\paragraph{LLM dependence in the pipeline.}
Canonicalisation, cuisine tagging, and the factor/mode label
generation all use Claude under deterministic decoding, and every
LLM-touched output is logged and inspectable.  The embeddings
themselves are LLM-free -- the skip-gram objective sees only walk
sequences over canonical ingredient and compound tokens -- so the
geometry we analyse is not directly conditioned on LLM judgements,
but the canonical vocabulary that defines its node set is.

\section{Conclusions}\label{sec:conclusions}

Computational gastronomy has moved from the descriptive flavour
network of \citet{ahn2011flavor}, through compound catalogues
(FlavorDB~\citep{garg2018flavordb}, FooDB~\citep{foodb2020}) and
integrated knowledge graphs
(FoodKG~\citep{haussmann2019foodkg}), to distributed-representation
food embeddings typified by
FlavorGraph~\citep{park2021flavorgraph}.  Epicure suggests the next
step is to expose the operators that act on such an embedding: a
300-D vector becomes useful to a chef when it is wrapped in
nearest-neighbour pairings, closest-mode lookup, and SLERP rotation
by a continuous angle, and when the inductive biases inside it are
exposed as named, controllable axes rather than hidden in the choice
of network.  Three openings extend the work directly: a continuous
mixing parameter at the walker that would turn the three siblings
into a parameterised family and let the chemistry-vs-recipe-context
trade-off be tuned rather than chosen; a richer set of operators
beyond single mode jumps -- intra-mode interpolation, multi-direction
blends, and constrained traversal (\emph{rotate toward Mediterranean
but stay in the dairy mode}); and cross-modal grounding through the
shared canonical vocabulary, so the SLERP operator can cross from
ingredient space into recipe-text, image, or sensory-descriptor space
on the same model.  More broadly, the methodological move of treating
the walk schema as the experimental variable applies to any future
fusion of culinary signals.  The next concrete artefact is a single
chef-facing interface that exposes all three controls -- model choice
(Cooc/Core/Chem), closest-mode lookup, and the SLERP angle -- in one
place; measuring what real users do with that interface is the next
empirical step.

\section*{Declaration of Generative AI Use}

This work used large language models in two capacities.
\textbf{Data pipeline:} Anthropic Claude Opus family models (internal
deployment IDs 4.6 and 4.7)~\citep{anthropic2026models,anthropic2026opus46}
performed all
ingredient classification under deterministic decoding
(temperature~0--0.1), including translation of non-English terms,
canonical-vocabulary construction, dedup adjudication, 1:1 matching
against USDA FoodData Central and FlavorDB, cuisine-marker tagging,
and generation of the sensory scores used as direction-quality
ground truth.  Google's \texttt{gemini-embedding-001}
endpoint~\citep{google2026geminiembeddingapi,lee2025geminiembedding}
was used to compute cosine
similarity between canonical-name candidates during one dedup stage.  All LLM outputs were
validated by rule-based post-processing or human review.
\textbf{Writing assistance:} Anthropic Claude Opus family models
(internal deployment IDs 4.6 and 4.7)~\citep{anthropic2026models}
were used for
drafting, editing, and code generation.  All scientific claims,
experimental design, and interpretations are the authors' own.

\bibliographystyle{plainnat}
\bibliography{references}

\end{document}